\documentclass[runningheads]{llncs}
\usepackage{amsmath,amsfonts,bm}

\def\eqref#1{equation~\ref{#1}}
\def\1{\bm{1}}

\DeclareMathAlphabet{\mathsfit}{\encodingdefault}{\sfdefault}{m}{sl}
\SetMathAlphabet{\mathsfit}{bold}{\encodingdefault}{\sfdefault}{bx}{n}

\usepackage{cite}
\usepackage{amssymb}
\usepackage{amsmath}
\usepackage{makecell}
\usepackage{booktabs}
\usepackage{algorithm}
\usepackage[noend]{algorithmic}
\usepackage{graphicx}
\usepackage{subfigure}
\usepackage{wrapfig}
\def\etal{{\em et al.\/}\,}

\begin{document}
\title{Dropout's Dream Land: Generalization from Learned Simulators to Reality}
\author{Zac Wellmer \and
James T. Kwok}
\institute{Department of Computer Science and Engineering\\
Hong Kong University of Science and Technology, Hong Kong \\
\email{\{zwwellmer,jamesk\}@cse.ust.hk}}
\maketitle              % typeset the header of the contribution
\begin{abstract}
A World Model is a generative model used to simulate an environment.
World Models have proven capable of learning spatial and temporal representations of Reinforcement Learning environments.
In some cases, a World Model offers an agent the opportunity to learn entirely inside of its own dream environment.
In this work we explore improving the generalization capabilities from dream environments 
to real environments
(Dream2Real).
We present a general approach to improve a controller's ability to transfer from a neural network dream environment to reality at little additional cost.
These improvements are gained by drawing on inspiration from Domain Randomization, where the basic idea is to randomize as much of a simulator as possible without fundamentally changing the task at hand.
Generally, Domain Randomization assumes access to a pre-built simulator with configurable parameters but oftentimes this is not available.
By training the World Model using dropout, the dream environment is capable of creating a nearly infinite number of \textit{different} dream environments.
Previous use cases of dropout either do not use dropout at inference time or averages the predictions generated by multiple sampled masks (Monte-Carlo Dropout).
Dropout's Dream Land leverages each unique mask to create a diverse set of dream environments. 
Our experimental results show that Dropout's Dream Land is an effective technique to bridge the reality gap between dream environments and reality.
Furthermore, we additionally perform an extensive set of ablation studies.\footnote{The code is available at \url{https://github.com/zacwellmer/DropoutsDreamLand}}
\end{abstract}

\section{Introduction}

Reinforcement learning~\cite{sutton2018reinforcement} (RL) has experienced a flurry of success in recent years,
from learning to play Atari~\cite{mnih2015human} to achieving grandmaster-level performance in StarCraft II~\cite{vinyals2019grandmaster}.
However, in all these examples, the target environment is a simulator that can be directly trained in.
Reinforcement learning is often not a practical solution without a simulator of the environment.

Sometimes the target environment is expensive, dangerous, or even impossible to interact with.
In these cases, the agent is trained in a simulated source environment.
Approaches that train an agent in a simulated environment with the hopes of generalization to the target environment experience a common problem referred to as the \textit{reality gap}~\cite{jakobi1995noise}.
One approach to bridge the reality gap is Domain Randomization~\cite{tobin2017domain}.
The basic idea is that an agent which can perform well in an ensemble of simulations will also generalize to the real environment~\cite{antonova2017reinforcement,tobin2017domain,mordatch2015ensemble,sadeghi2016cad2rl}.
The ensemble of simulations is generally created by randomizing as much of the simulator as possible without fundamentally changing the task at hand.
Unfortunately, this approach is only applicable when a simulator is provided and the simulator is configurable.

A recently growing field, World Models~\cite{ha2018recurrent}, focuses on the side of this problem when the simulation does not exist.
World Models offer a general framework for optimizing controllers directly in \textit{learned} simulated environments.
The learned dynamics model can be viewed as the agent's dream environment.
This is an interesting area because access to a learned dynamics model removes the need for an agent to train in the target environment.
Some related approaches~\cite{Kaiser2020Model,hafner2019learning,Hafner2020Dream,sekar2020planning,sutton1990integrated,kurutach2018model} focus on an adjacent problem which allows the controller to continually interact with the target environment.

Despite the recent improvements~\cite{Kaiser2020Model,hafner2019learning,sekar2020planning,kim2020learning,Hafner2020Dream} of World Models, little has been done to address the issue that World Models are susceptible to the reality gap.
The learned dream environment can be viewed as the source domain and the true environment as the target domain.
Whenever there are discrepancies between the source and target domains the reality gap can cause problems.
Even though World Models suffer from the reality gap, none of the Domain Randomization approaches are directly applicable because the dream environment does not have easily configurable parameters.

In this work we present Dropout's Dream Land (DDL), a simple approach to bridge the reality gap from learned dream environments to reality (Dream2Real).
Dropout's Dream Land was inspired by the first principles of domain
randomization, namely, train a controller on a large set of \textit{different} simulators which all adhere to the fundamental task of the target environment.
We are able to generate a nearly infinite number of different simulators via the insight that dropout~\cite{srivastava2014dropout} can be understood as learning an ensemble of neural networks~\cite{baldi2013understanding}.

Our empirical results demonstrate that Dropout's Dream Land is an effective technique to cross the Dream2Real gap and offers improvements over baseline
approaches~\cite{ha2018recurrent,kim2020learning}.
Furthermore, we perform an extensive set of ablation studies which indicate the source of generalization improvements, requirements for the method to work, and when the method is most useful.

\section{Related Works}

\subsection{Dropout}
Dropout~\cite{srivastava2014dropout} was introduced as a regularization technique for feedforward and convolutional neural networks.
In its most general form, each unit is dropped with a probability $p$ during the training process.
During training weights are scaled by $\frac{1}{1-p}$. 
Weight scaling ensures that for any hidden unit the \textit{expected} output is the same as the actual output at test time~\cite{srivastava2014dropout}.
Recurrent neural networks (RNNs) initially had issues benefiting from dropout. 
Zaremba \etal
\cite{zaremba2014recurrent} 
suggests not to apply dropout to the hidden state units of the RNN cell.
Gal \etal
\cite{gal2016theoretically} shortly after show that the mask can also be
applied to the hidden state units, but the mask must be fixed across the sequence during training.

In this work, we follow the dropout approach from \cite{gal2016theoretically} when training the RNN.
More formally, for each sequence, the Boolean masks
$\mathbf{m}_{xi}$, $\mathbf{m}_{xf}$, $\mathbf{m}_{xw}$, $\mathbf{m}_{xo}$,
$\mathbf{m}_{hi}$, $\mathbf{m}_{hf}$, $\mathbf{m}_{hw}$, and $\mathbf{m}_{ho}$
are sampled, then used in the following LSTM update:
\begin{align}
    \mathbf{i}_t & = \mathbf{W}_{xi} (\mathbf{x}_t \odot \mathbf{m}_{xi}) +
	 \mathbf{W}_{hi}(\mathbf{h}_{t-1}\odot \mathbf{m}_{hi}) + \mathbf{b}_i, \label{eq:i_drop_gate} \\
    \mathbf{f}_t & = \mathbf{W}_{xf} (\mathbf{x}_t \odot \mathbf{m}_{xf}) +
	 \mathbf{W}_{hf}(\mathbf{h}_{t-1}\odot \mathbf{m}_{hf}) + \mathbf{b}_f, \label{eq:f_drop_gate} \\
    \mathbf{w}_t & = \mathbf{W}_{xw} (\mathbf{x}_t \odot \mathbf{m}_{xw}) +
	 \mathbf{W}_{hw}(\mathbf{h}_{t-1}\odot \mathbf{m}_{hw}) + \mathbf{b}_w, \label{eq:w_drop} \\
    \mathbf{o}_t & = \mathbf{W}_{xo} (\mathbf{x}_t \odot \mathbf{m}_{xo}) +
	 \mathbf{W}_{ho}(\mathbf{h}_{t-1}\odot \mathbf{m}_{ho}) + \mathbf{b}_o, \label{eq:o_drop} \\
	\mathbf{c}_t & = \sigma(\mathbf{i}_t) \odot \tanh(\mathbf{w}_t) + \sigma(\mathbf{f}_t) \odot \mathbf{c}_{t-1}, \label{eq:c} \\
    \mathbf{h}_t & = \sigma(\mathbf{o}_t) \odot \tanh(\mathbf{c}_t), \label{eq:h}
\end{align}
where $\mathbf{x}_t$, $\mathbf{h}_t$, and $\mathbf{c}_t$ are the input, hidden
state, and cell state, respectively, $\mathbf{W}_{xi}$, $\mathbf{W}_{xf}$, $\mathbf{W}_{xw}$, $\mathbf{W}_{xo} \in \mathbb{R}^{d \times r}$
$\mathbf{W}_{hi}$, $\mathbf{W}_{hf}$, $\mathbf{W}_{hw}$, $\mathbf{W}_{ho} \in \mathbb{R}^{d \times d}$ are the LSTM weight matrices, and $\mathbf{b}_{i}$, $\mathbf{b}_{f}$, $\mathbf{b}_{w}$, $\mathbf{b}_{o} \in \mathbb{R}^{d}$ are the LSTM biases.
The masks are fixed for the entire sequence, but may differ between sequences in the mini-batch.

Monte-Carlo (MC) Dropout~\cite{gal2016dropout} runs multiple forward passes with independently sampled masks.
In related works~\cite{kahn2017uncertainty}, Monte-Carlo (MC) Dropout~\cite{gal2016dropout} has been used to approximate the mean and variance of output predictions from an ensemble.
We emphasize that Dropout's Dream Land does not use MC Dropout. 
Details are in Section~\ref{ssec:dropout_env}.

\subsection{Domain Randomization}\label{ssec:DR}
The goal of Domain Randomization~\cite{tobin2017domain,sadeghi2016cad2rl} is to create many different versions of the dynamics model with the hope that a policy generalizing to all versions of the dynamics model will do well on the true environment.
Figure~\ref{fig:DR} illustrates many simulated environments ($\hat{e}^j$) overlapping with the actual environment ($e^*$).
Simulated environments are often far cheaper to operate in than the actual
environment. Hence, it is desirable to be able to perform the majority of interactions in the simulated environments.

Randomization has been applied on observations (e.g., lighting, textures) to perform robotic grasping~\cite{tobin2017domain} and collision avoidance of drones~\cite{sadeghi2016cad2rl}.
Randomization has also proven useful when applied to the underlying dynamics of simulators~\cite{peng2018sim}.
Often, both the observations and simulation dynamics are
randomized~\cite{andrychowicz2020learning}.

Domain randomization generally uses some pre-existing simulator which then
injects randomness into specific aspects of the simulator (e.g., color textures, friction coefficients).
Each of the simulated environments in Figure~\ref{fig:DR} can be thought of as a noisy sample of the pre-existing simulator.
To the best of our knowledge, Domain Randomization has yet to be applied to entirely learned simulators.

\subsection{World Models}\label{ssec:wm}

\begin{figure*}[t]
    \begin{minipage}{0.52\linewidth}
        \begin{algorithm}[H]
          \caption{World Models: Training in dreams.}\label{alg:wm_learning}
          \begin{algorithmic}[1]
            \STATE Initialize parameters of $V$, $M$, and $C$
            \STATE Collect $N$ trajectories $\mathbf{o}$, $d$, and $\mathbf{a}$ from $e^*$
            \STATE Optimize $V$ on observations $\mathbf{o}$
            \STATE Generate embeddings $\mathbf{z}$ for $\mathbf{o}$ with $V$
            \STATE Optimize $M$ on $\mathbf{z}$ and $d$ \label{line:m_train}
            \STATE Generate dream environment $\hat{e}$ from $M$
            \FOR{iteration=1, 2, $\dotsc$}
                \STATE Optimize $C$ via interactions with $\hat{e}$ \label{line:c_train}
            \ENDFOR
          \end{algorithmic}
        \end{algorithm}
    \end{minipage}
    \hfill
    \begin{minipage}{0.45\linewidth}
        \centering
        \includegraphics[height=3cm]{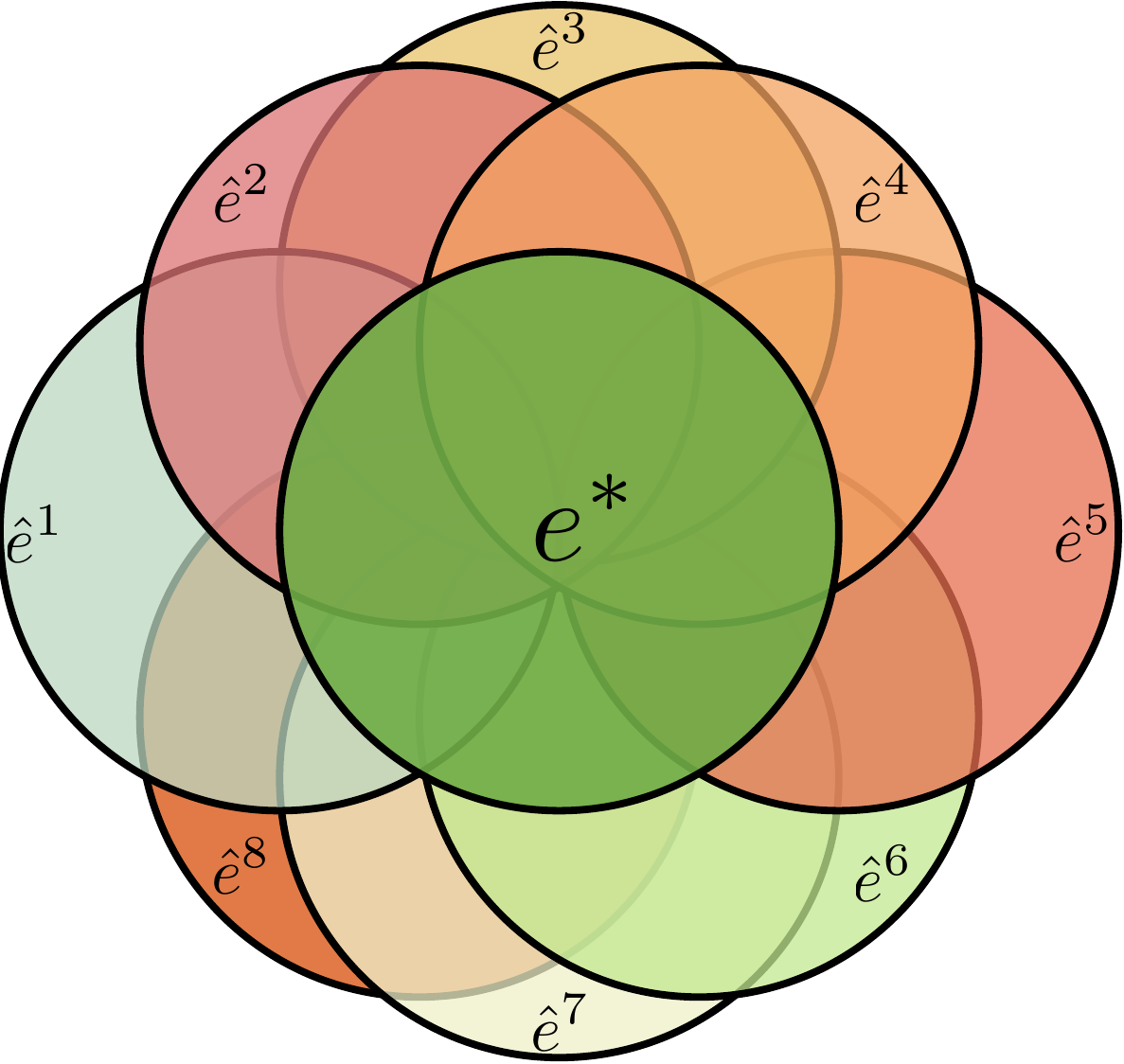}
        \caption{$e^*$ is the actual environment, and $\hat{e}^j$'s are randomized variants of the simulated environment.}
        \label{fig:DR}
    \end{minipage}
\end{figure*}

The world model
\cite{ha2018recurrent}
has three modules trained separately:
(i) vision module ($V$);
(ii) dynamics module ($M$); and
(iii) controller ($C$).
A high-level view is shown in Algorithm~\ref{alg:wm_learning}.
The vision module ($V$) is a variational autoencoder (VAE)~\cite{kingma2013auto}, which maps an image observation ($\mathbf{o}$) to a lower-dimensional representation $\mathbf{z} \in \mathbb{R}^n$.

The dynamics model ($M$) is a mixture density network recurrent neural network
(MDN-RNN)~\cite{ha2018recurrent,graves2013generating}.
The MDN-RNN models the dynamics of the environment,
so modifying the parameters changes the dynamics of the learned simulated environment.
It is implemented as an LSTM followed by a fully-connected layer outputting parameters for a Gaussian mixture model with $k$ components.
Each
feature has $k$ different
$\pi$
parameters
for the logits of multinomial distribution, and
$(\mu, \sigma)$
parameters for the $k$ components in the Gaussian mixture.
At each timestep,
the MDN-RNN takes in the state $\mathbf{z}$ and action $\mathbf{a}$ as inputs and
predicts
$\bm{\pi}, \bm{\mu}$, $\bm{\sigma}$.
To draw a sample from the MDN-RNN, we first sample the multinomial distribution parameterized
by $\bm{\pi}$, which indexes which of the $k$ normal distributions
in the Gaussian mixture to sample from. This is then repeated for each of the $n$ features.
Depending on the experiments,
Ha and Schmidhuber \cite{ha2018recurrent} also include an auxiliary head to the LSTM which predicts whether the episode terminates ($d$).

The controller ($C$) is responsible for deciding what actions to take.
It takes features produced by the encoder $V$ and dynamics model $M$ as input (not the raw observations).
The simple controller is a single-layer model which uses an evolutionary algorithm (CMA-ES \cite{hansen2001completely}) to find its parameters.
Depending on the problem setting, the controller ($C$) can either be optimized directly on the target environment ($e^*$) or on the dream environment ($\hat{e}$).
This paper is focused on the case of optimizing exclusively in the dream environment. 

\section{Dropout's Dream Land}\label{sec:dream_land}

In this work we introduce Dropout's Dream Land (DDL).
Dropout's Dream Land is the first work to offer a strategy to bridge the \textit{reality gap} between learned neural network dynamics models and reality.
Traditional Domain Randomization generates many \textit{different} dynamics models by randomizing configurable parameters of a given simulation.
This approach does not apply to neural network dynamics models because they
generally do not have configurable parameters (such as textures and friction coefficients).
In Dropout's Dream Land, the controller can interact with billions\footnote{In practice we are bounded by the total number of steps instead of every possible environment.}
of dream environments, whereas previous works~\cite{ha2018recurrent,kim2020learning} only use one dream environment.
A naive way to go about this would be to train a population of neural network world models.
However, this would be computationally expensive.

To keep the computational cost low, we go about this by applying dropout to the dynamics model in order to form different dynamics models.
Crucially, dropout is applied at \textbf{both} training and inference of the dynamics model $M$.
Each unique dropout mask applied to $M$ can be viewed as a different environment.
Similar to the spirit of Domain Randomization, an agent is expected to perform well in the real environment if it can perform well in a variety of simulated environments.

\subsection{Learning the Dream Environment}\label{ssec:build}

\begin{figure*}[t]
    \begin{minipage}{0.49\linewidth}
        \vspace{-.75cm}
        \includegraphics[width=\linewidth]{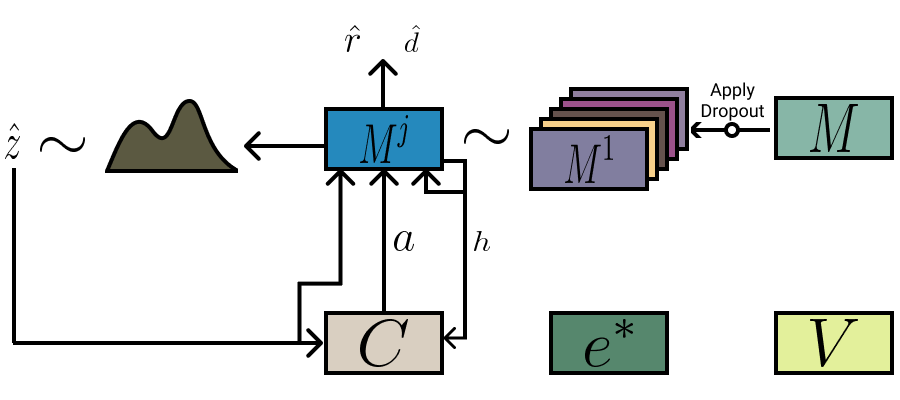}
        \caption{Interactions with the dream environment. A dropout mask is sampled at every step yielding a new $M^j$.}
        \label{fig:dropout_env}
    \end{minipage}
    \hfill
    \begin{minipage}{0.49\textwidth}
        \includegraphics[width=\linewidth]{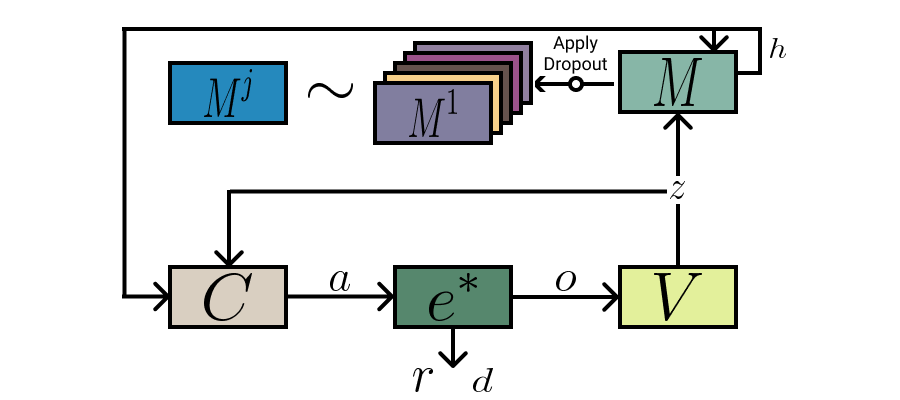}
        \vspace{-.75cm}
        \caption{Interactions with the real environment. The controller being optimized only interacts with the real environment during the final testing phase.}
        \label{fig:env}
    \end{minipage}
\end{figure*}

The Dropout's Dream Land environments are built around the dynamics model $M$.
The controller interactions 
during training 
are described by Figure~\ref{fig:dropout_env}, in which
$\hat{r}$, $\hat{d}$, and $\mathbf{\hat{z}}$ are generated entirely by $M$.
In this work, $M$ is an LSTM where $\mathbf{x}= [\mathbf{z}^\top, \mathbf{a}^\top]^\top$ from equations~(\ref{eq:i_drop_gate})-(\ref{eq:o_drop}).
The LSTM is followed by multiple heads for predictions of the latent state ($\mathbf{\hat{z}}$),
reward ($\hat{r}$)
and
termination ($\hat{d}$).
The reward and termination heads are simple fully-connected layers.
Latent state prediction is done with a
MDN-RNN~\cite{ha2018recurrent,graves2013generating}, but this could be
replaced by any other neural network that supports dropout (e.g., GameGAN~\cite{kim2020learning}).

\subsubsection{Loss Function}

The dynamics model $M$ jointly optimizes all three heads. The loss of a single transition is defined as:
\begin{equation} \label{eq:M_loss}
\mathcal{L}^M =
\mathcal{L}^z +
\alpha_r \mathcal{L}^r +
\alpha_d \mathcal{L}^d.
\end{equation}
Here,
$\mathcal{L}^z = -\sum^n_{i=1} \log (\sum^k_{j=1} \hat{\pi}_{i,j}
\mathcal{N}(z_i | \hat{\mu}_{i,j}, \hat{\sigma}_{i,j}^2))$
is a mixture density loss for the latent state predictions,
where $n$ is the size of the latent feature vector $z$, $\hat{\pi}_{i,j}$ is the
$j$th component's probability for the $i$th feature, $\hat{\mu}_{i,j}, \hat{\sigma}_{i,j}$ are the corresponding mean and standard deviation.
$\mathcal{L}^r = (r - \hat{r})^2$
is the square loss on rewards,  where
$r$ and $\hat{r}$ are the true and estimated rewards, respectively.
$\mathcal{L}^d = -d\log(\hat{d}) - (1-d)\log(1-\hat{d})$
is
the cross-entropy loss
for termination prediction,
where $d$ and $\hat{d}$ are the true and estimated probabilities of the episode ending, respectively.
Constants $\alpha_d$ and $\alpha_r$
in
(\ref{eq:M_loss})
are for trading off importance of the termination and reward objectives.
The loss ($\mathcal{L}^M$) is aggregated over each sequence and averaged across the mini-batch.

\subsubsection{Training Dynamics Model $M$ with Dropout}\label{sssec:rnn_dropout}

At training time of $M$ (Algorithm~\ref{alg:wm_learning}, Line~\ref{line:m_train}), we apply dropout~\cite{gal2016theoretically} to the LSTM to simulate different random environments. 
For each input and hidden unit, we first sample a Boolean indicator with probability $p_\text{train}$.
If the indicator is 1, the corresponding input/hidden unit is masked.
Masks $\mathbf{m}_{xi}$, $\mathbf{m}_{xf}$, $\mathbf{m}_{xw}$, $\mathbf{m}_{xo}$,
$\mathbf{m}_{hi}$, $\mathbf{m}_{hf}$, $\mathbf{m}_{hw}$, and $\mathbf{m}_{ho}$
are sampled independently (Equations~(\ref{eq:i_drop_gate})-(\ref{eq:o_drop})).
When training the RNN, each mini-batch contains multiple sequences.
Each sequence uses an independently sampled dropout mask.
We fix the dropout mask for the entire sequence as this was previously found to
be critically important~\cite{gal2016theoretically}.

Training the RNN with many different dropout masks is critical in order to generate multiple different dynamics models.
At the core of Domain Randomization is the requirement that the randomizations do not fundamentally change the task.
This constraint is violated if we do not train the RNN with dropout but apply
dropout at inference (explored further empirically in Section~\ref{ssec:dropout_abl}).
After optimizing the dynamics model $M$, we can use it to construct dream environments (Section~\ref{ssec:dropout_env}) for controller training (Section~\ref{ssec:controller}).

In this work, we never sample masks to apply to the action ($\mathbf{a}$).
We do not zero out the action because in some environments this could imply the
agent taking an action (e.g., moving to the left).
This design choice could be changed depending on the environment, for example, when a zero’d action corresponds to a no-op or a sticky action.

\subsection{Interacting with Dropout's Dream Land} \label{ssec:dropout_env}

Interactions with the dream environment (Algorithm~\ref{alg:wm_learning},  Line~\ref{line:c_train}) can be characterized as training time for the controller ($C$) and inference time of the dynamics model ($M$).
An episode begins by generating the initial latent state vector $\mathbf{\hat{z}}$ by either sampling from a standard normal distribution or sampling from the starting points of the observed trajectories used to train $M$ \cite{ha2018recurrent}.
The hidden cell ($\mathbf{c}$) and state ($\mathbf{h}$) vectors are initialized with zeros.

The controller ($C$) decides the action to take based on $\mathbf{\hat{z}}$ and $\mathbf{h}$.
In Figure~\ref{fig:dropout_env}, the controller also observes $\hat{r}$ and
$\hat{d}$, but these are exclusively used for the optimization process of the controller.
The controller then performs an action $\mathbf{a}$ on a dream environment.

A new dropout mask is sampled (with probability $p_\text{infer}$) and applied to $M$.
We refer to the masked dynamics model as $M^j$ and the corresponding Dropout's Dream Land environment as $\hat{e}^j$.
The current latent state $\mathbf{\hat{z}}$ and action $\mathbf{a}$ are concatenated,
and passed to $M^j$ to perform a forward pass.
The episode terminates based on a sample from a Bernoulli distribution parameterized by $\hat{d}$.
The dream environment then outputs the latent state, LSTM's hidden state, reward, and whether the episode terminates.

It is crucial to apply dropout at inference time (of the dynamics model $M$) in order to create \textit{different} versions of the dream environment for the controller $C$.
Our experiments (Sections~\ref{ssec:dropout_ret_exp} and \ref{ssec:dropout_abl}) consider an extensive set of ablation studies as to how and when dropout should be applied.

\subsubsection{Dropout's Dream Land is not Monte-Carlo Dropout}
The only work we are aware of that applies dropout at inference time is
Monte-Carlo Dropout~\cite{gal2016theoretically}.
In Section~\ref{ssec:baselines} we include a Monte-Carlo Dropout World Model baseline because DDL can easily be misinterpreted as an application of Monte-Carlo Dropout.
This baseline passes the
expected hidden
($\bm{\tilde{h}}_t$) and cell ($\bm{\tilde{c}}_t$) state to the next time-step,
in which
the expectation is over dropout masks from Equations~(\ref{eq:i_drop_gate})-(\ref{eq:o_drop}). 
In practice we follow a similar approach to previous
work~\cite{gal2016theoretically} and approximate the expectation by performing
multiple forward passes (each forward pass samples a new dropout mask), and averages the results.
At each step, the expected Mixture Model parameters ($\bm{\tilde{\pi}}$, $\bm{\tilde{\mu}}$, $\bm{\tilde{\sigma}}$), reward ($\tilde{r}$), and termination ($\tilde{d}$) are used. 
Maximizing expected returns from the Monte-Carlo Dropout World Model is equivalent to maximizing expected returns on a \textit{single} dream environment, the average dynamics model. 
On the other hand,
the purpose of DDL's approach to dropout is to generate many \textit{different} versions of the dynamics model.
More explicitly, the controller is trained to maximize expected returns across
many different dynamics models in the ensemble, as opposed to maximizing expected returns on the ensemble average.

Dropout has also traditionally been used as a model regularizer.
Dropout as a model regularizer is only applied at training time but not at inference time.
In this work, this approach would regularize the dynamics model $M$.
The usual trade-off is lower test loss at the cost of higher training loss~\cite{srivastava2014dropout,gal2016theoretically}.
However, DDL's ultimate goal is not to lower test loss of the World Model ($M$).
The ultimate goal is providing dream environments to a controller so that the optimal policy in Dropout's Dream Land also maximizes expected returns in the target environment ($e^*$).

\subsection{Training the Controller}\label{ssec:controller}
\subsubsection{Training with CMA-ES}\label{sssec:cma}

We follow the same controller optimization procedure as was done in World Models~\cite{ha2018recurrent} and GameGAN~\cite{kim2020learning} on their DoomTakeCover experiments.
We train the controller with CMA-ES \cite{hansen2001completely}.
At every generation CMA-ES~\cite{hansen2001completely} spawns a population (of size $N_\text{pop}$) of agents.
Each agent in the population reports their mean
returns on a set of $N_\text{trials}$ episodes generated in
the dream environments.
As controllers in the population do not share
a dream environment, the probability of controllers interacting with the same sequence of dropout masks is vanishingly small.
Let $N_\text{max\_ep\_len}$ be the maximum number of steps in an episode.
In a single CMA-ES iteration, the population as a whole can interact with $N_\text{pop} \times N_\text{trials} \times N_\text{max\_ep\_len}$ \textit{different} environments.
In our experiments, $N_\text{pop} = 64$, $N_\text{trials}= 16$, and $N_\text{max\_ep\_len}$ is $1000$ for CarRacing and $2100$ for DoomTakeCover.
This potentially results in $>1,000,000$ different environments at each generation.

\subsubsection{Dream Leader Board}\label{sssec:leaderboard}
After every fixed number of generations ($25$ in our experiments), the best controller in the population (which received the highest average returns across its respective $N_\text{trials}$ episodes) is selected for evaluation~\cite{ha2018recurrent,kim2020learning}.
This controller is evaluated for another
$N_\text{pop} \times N_\text{trials}$ episodes in the Dropout's Dream Land environments.
The controller's mean across $N_\text{pop} \times N_\text{trials}$ trials is logged to the Dream Leader Board.
After $2000$ generations, the controller at the top of the Dream Leader Board is evaluated in the real environment.

\subsubsection{Interacting with the Real Environment}\label{sssec:target}

In Figure~\ref{fig:env} we illustrate the controller's interaction with the real
target
environment ($e^*$).
Interactions with $e^*$ do not apply dropout to the input or hidden units of $M$.
The controller only interacts with the target environment during testing.
These interactions are never used to modify parameters of the controller.
At test time $r$, $d$, and $o$ are generated by $e^*$, and $\mathbf{z}$ is the embedding of $o$ from the VAE ($V$).
The only use of $M$ when interacting with the target environment is producing $\mathbf{h}$ as a feature for the controller.

\section{Experiments}\label{sec:exp}
Broadly speaking, our experiments are focused on either evaluating the dynamics model ($M$) or the controller ($C$).
Architecture details of $V$, $M$, and $C$ are in Appendix~\ref{A:arch}.
Experiments are performed on the DoomTakeCover-v0~\cite{DoomTakeCover-v0} and
CarRacing-v0~\cite{CarRacing-v0} environments from OpenAI
Gym~\cite{brockman2016openai}.
These have also been used in related works~\cite{kim2020learning,ha2018recurrent}.
Even though both baseline target environments are simulators we still consider
this ``reality" because we do not leverage knowledge about the simulator mechanics to learn the source environment ($M$). 

Quality of the dynamics model is evaluated against a training and testing set of trajectories (described below).
Quality of the controller is measured by returns in the target environments.
For all experiments the controller is trained exclusively in the dream
environment (Section~\ref{ssec:dropout_env}) for 2,000 generations.
The controller only interacts with the target environments for testing (Section~\ref{sssec:target}). 
The target environment is never used to update parameters of the controller.
Means and standard deviations of returns achieved by the best controller (Section~\ref{sssec:leaderboard}) in the target environment are reported based on $100$ trials for CarRacing and $1000$ trials for DoomTakeCover.\footnote{100 trials are used for the baselines GameGAN and Action-LSTM.}

\textbf{DoomTakeCover Environment}
DoomTakeCover is a control task in which the goal is to dodge fireballs for as long as possible.
The controller receives a reward of +1 for every step it is alive.
The maximum number of frames is limited to $2100$.

For all tasks on this environment, we collect a training set of $10,000$ trajectories and a test set of $100$ trajectories.
A trajectory is a sequence of state ($\mathbf{z}$), action ($\mathbf{a}$), reward ($r$), and termination ($d$) tuples.
Both datasets are generated according to a random policy.
Following the same convention as World Models~\cite{ha2018recurrent}, on the DoomTakeCover environment we concatenate $\bm{z}$, $\bm{h}$, and $\bm{c}$ as input to the controller.
In (\ref{eq:M_loss}), we set $\alpha_d=1$ and $\alpha_r=0$ because the Doom reward function is determined entirely based on whether the controller lives or dies.

\textbf{CarRacing Environment}
CarRacing is a continuous control task to learn from pixels.
The race track is split up into ``tiles".
The goal is to make it all the way around the track (i.e., crossing every tile).
We terminate an episode when all tiles are crossed or when the number of steps
exceeds 1,000.
Let $N_\text{tiles}$ be the total number of tiles.
The simulator~\cite{CarRacing-v0} defines the reward
$r_t$
at each timestep as
$\frac{100}{N_\text{tiles}} -0.1$ if a new tile is crossed, and $-0.1$ otherwise.
The number of tiles is not explicitly set by the simulator.
We generated 10,000 tracks and
observed that the number of tiles
in the track appears to follow a normal
distribution with mean $289$.
To simplify the reward function, we fix $N_\text{tiles}$ to
289 in the randomly generated tracks, and call the modified environment CarRacingFixedN.

For all tasks on this environment, the training set contains $5,000$ trajectories and the test set contains $100$ trajectories.
Both datasets are collected by following an expert policy with probability $0.9$, and a random policy with probability $0.1$.
The expert policy was trained directly on the CarRacing environment and received an average return of $885 \pm 63$ across $100$ trials.
In comparison, the performance of the random policy is $-53 \pm 41$.
This is similar to the setup in GameGAN \cite{kim2020learning}
on the Pacman%\footnote{Proprietary environment that we do not have access to.}
environment which also used an expert policy.
For this environment, we set $\alpha_d=\alpha_r=1$
in (\ref{eq:M_loss}).

\subsection{Comparison with Baselines}\label{ssec:baselines}
Dropout's Dream Land (DDL) is compared against World Models (WM), Monte-Carlo
Dropout World Models (MCD-WM), and a uniform random policy on the CarRacing and
DoomTakeCover environments.
The Monte-Carlo Dropout World Models baseline uses $p_\text{train}=0.05$,
$p_\text{infer}=0.1$, and 10 samples.
On the Doom environment, we also compare with GameGAN~\cite{kim2020learning} and
Action-LSTM~\cite{chiappa2017recurrent}\footnote{Results on GameGAN and
Action-LSTM returns are from~\cite{kim2020learning}.}.
All controllers are trained entirely in dream environments. 

Results on the target environments are in Tables~\ref{tab:doom_baseline}
and \ref{tab:car_baseline}.
The CarRacing results appear different from those found in World Models~\cite{ha2018recurrent} because we are not performing the same experiment. 
In this paper, we train the controller entirely in the dream environment and only interact with the target environment during testing. 
In World Models~\cite{ha2018recurrent}, the controller was trained directly in the CarRacing environment.

\begin{figure*}[t]
    \begin{minipage}{0.45\linewidth}
        \begin{table}[H]
            \caption{Returns from baseline methods and DDL  ($p_\text{train}=0.05$ and $p_\text{infer}=0.1$) on the DoomTakeCover environment.}
            \centering
            \begin{tabular}{lr}
                \toprule
                & DoomTakeCover \\
                \midrule
                random policy & $210 \pm 108$ \\ 
                GameGAN & $765 \pm 482$ \\ 
                Action-LSTM & $280 \pm 104$ \\ 
                WM & $ 849 \pm 499$ \\ 
                MCD-WM & $798 \pm 464$ \\
                DDL & $ \mathbf{933 \pm 552}$ \\
                \bottomrule
            \end{tabular}
            \label{tab:doom_baseline}
        \end{table}
    \end{minipage}
    \hfill
    \begin{minipage}{0.53\linewidth}
        \begin{table}[H]
            \caption{Returns from baseline methods and DDL ($p_\text{train}=0.05$ and $p_\text{infer}=0.1$) on the CarRacingFixedN 
and the original CarRacing environments.}
            \centering
            \begin{tabular}{ccc}
                \toprule
                & CarRacingFixedN & CarRacing \\
                \midrule 
                random policy & $-50 \pm 38$ & $-53 \pm 41$ \\
                WM & $399 \pm 135$ & $388 \pm 157 $ \\
                MCD-WM & $-56 \pm 31$ & $-53 \pm 32$ \\
                DDL & $\mathbf{625 \pm 289}$ & $\mathbf{610 \pm 267}$ \\
                \addlinespace[.7cm]
                \bottomrule
            \end{tabular}
            \label{tab:car_baseline}
        \end{table}
    \end{minipage}
\end{figure*}

In Tables~\ref{tab:doom_baseline} and~\ref{tab:car_baseline}, we observe that DDL offers performance improvements over all the baseline approaches in the target environments.
We suspect this is because the WM dream environments were easier for the controller to exploit errors between the simulator and reality.
Forcing the controller to succeed in many different dropout environments makes it difficult to exploit discrepancies between the dream environment and reality.
This leads us to the conclusion that forcing the controller to succeed in many different dropout environments is an effective technique to cross the Dream2Real gap.

The DoomTakeCover returns in the target environment 
as reported by the temperature-regulated variant\footnote{We were unable to reproduce the temperature results
in~\cite{ha2018recurrent}.}
in~\cite{ha2018recurrent}
are higher than the returns  we obtain
from DDL, which does not use temperature.
However, we emphasize that adjusting temperature is only useful for a limited set of dynamics models.
For example, it would not be straightforward to apply temperature to any dynamics
model which does not produce a probability density function (e.g., GameGAN);
whereas the DDL approach of generating many \textit{different} dynamics models is
useful to any learned neural network dynamics model.
Moreover, even though the temperature-regulated variant increases uncertainty of the dream
environment, it is still only capable of creating \textit{one} dream environment.

\subsection{Inference Dropout and Dream2Real Generalization}\label{ssec:dropout_ret_exp}

\begin{table}[t]
    \caption{RNN's loss with and without dropout ($p_\text{train}=0.05$ and $p_\text{infer}=0$) during training.}
    \label{tab:rnn_loss}
    \centering
    \begin{tabular}{ccccc}
        \toprule
        \multicolumn{1}{c}{} & \multicolumn{2}{c}{DoomTakeCover} & \multicolumn{2}{c}{CarRacingFixedN} \\
        \cmidrule{2-5}
        & training loss & test loss & training loss & test loss\\
        \hline
        without dropout & $0.89$ & $\bm{0.91}$ & $2.36$ & $\bm{3.10}$ \\
        \hline
        with dropout & $0.93$ & $\bm{0.91}$ & $3.19$ &  $3.57$ \\
        \bottomrule
    \end{tabular}
\end{table}

\begin{figure}[t]
    \centering
    \subfigure[DoomTakeCover.]{\includegraphics[width=0.48\linewidth]{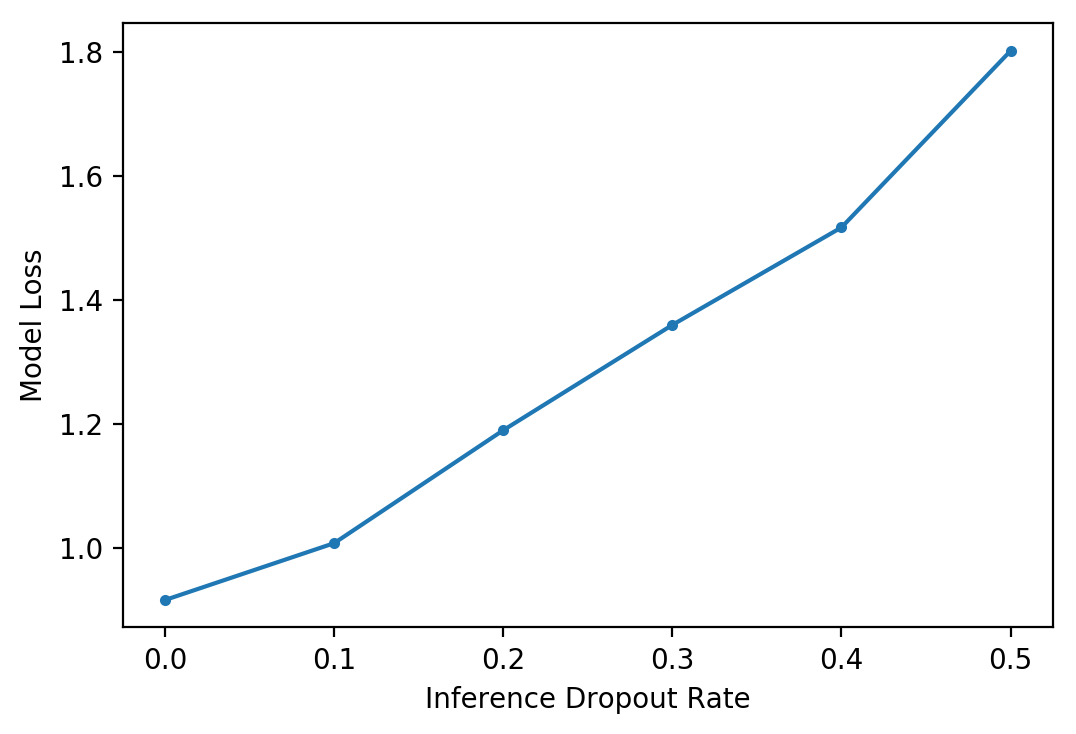}}
    \hfill
    \subfigure[CarRacingFixedN.]{\includegraphics[width=0.48\linewidth]{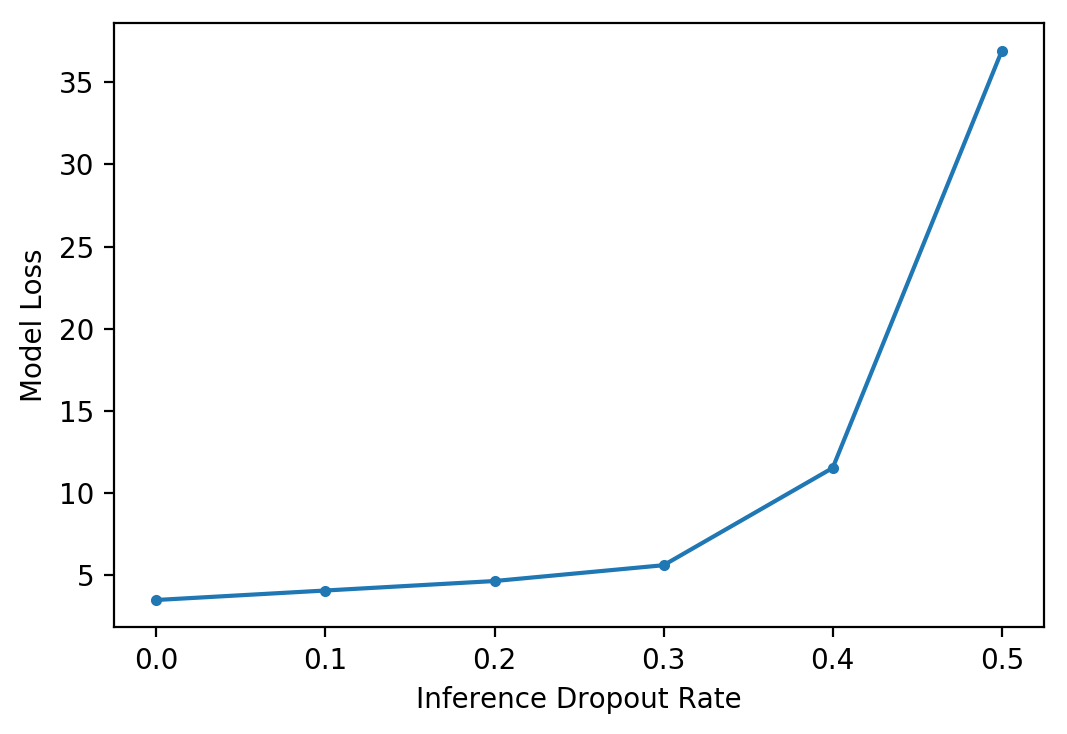}}
    \caption{Loss of DDL dynamics model ($p_\text{train}=0.05$) at different inference dropout rates.}
    \label{fig:model_loss}
    \centering
    \subfigure[DoomTakeCover.]{\includegraphics[width=0.48\linewidth]{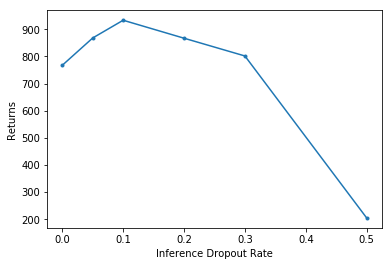}}
    \hfill
    \subfigure[CarRacingFixedN.]{\includegraphics[width=0.48\linewidth]{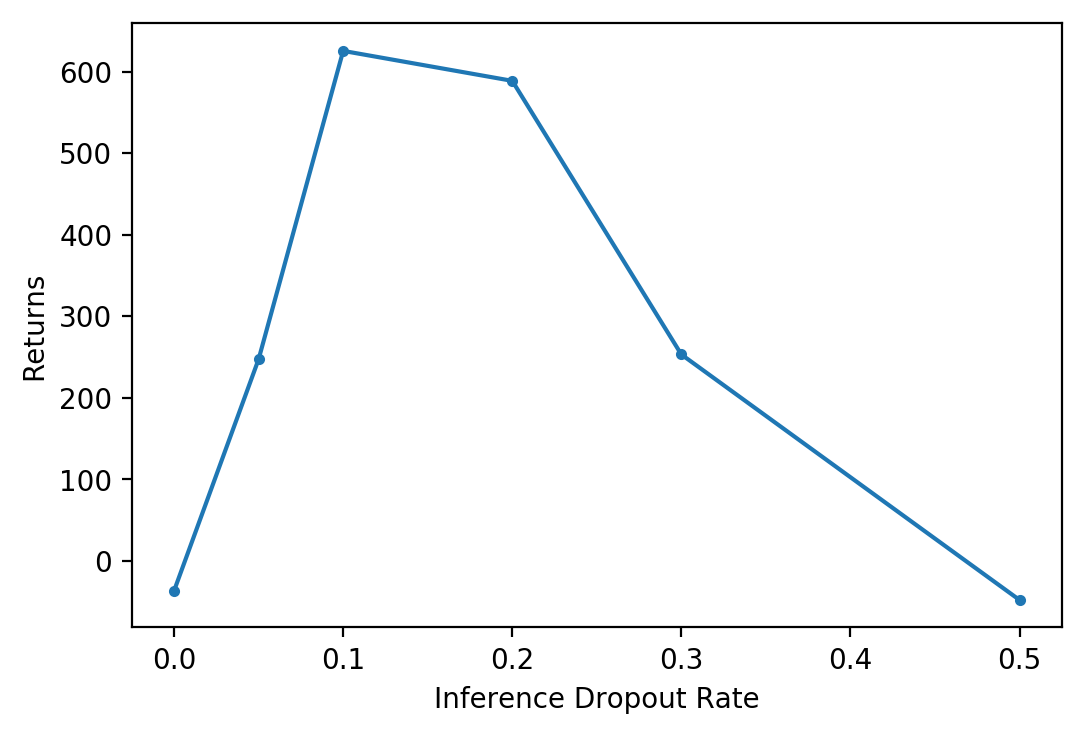}}
    \caption{DDL ($p_\text{train}=0.05$) returns at different inference dropout
	 rates in the target environments.}
    \label{fig:dropout_returns}
\end{figure}

In this experiment, we study the effects of dropout on the World Model.
First, we evaluate the relationship between dropout and World Model accuracies.
Second, we evaluate the relationship between dropout and generalization from the World Model to the target environment.
Model loss is measured by the loss in (\ref{eq:M_loss}) on the test sets.
Returns in the target environment are reported based on the best controller (Section~\ref{sssec:leaderboard}) trained with varying levels of inference dropout.
The same training and test sets described at the beginning of Section~\ref{sec:exp} are used.

Standard use cases of dropout generally observe a larger training loss but lower test loss relative to the same model trained without dropout~\cite{srivastava2014dropout,gal2016dropout}.
In Table~\ref{tab:rnn_loss}, we do not observe any immediate performance improvements of the World Model trained with dropout ($p_\text{train}=0.05$ and $p_\text{infer}=0$).
In fact, we observe worse results on the test set.
The poor performance of both DDL RNNs (Table~\ref{tab:rnn_loss})
indicates a clear conclusion about the results from Tables~\ref{tab:doom_baseline} and~\ref{tab:car_baseline}.
The improved performance of DDL relative to World Models comes from forcing the controller to operate in many different environments and not from a single more accurate dynamics model $M$.

Next we take a World Model trained with dropout and evaluate the model loss on a test set across varying levels of inference dropout ($p_\text{infer}$).
As expected, in Figure~\ref{fig:model_loss} we observe that as the inference dropout rate is increased the model loss increases.
In Figure~\ref{fig:dropout_returns} we observe that increasing the inference dropout rate improves generalization to the target environment.
We believe that the boost in returns on the target environments comes from an increase in capacity to distort the dynamics model.
Figures~\ref{fig:model_loss} and~\ref{fig:dropout_returns} suggest that we can sacrifice accuracy of the dream environments to better cross the Dream2Real gap between dream and target environments.
However, this should only be useful up to the point where \textit{the task at hand is fundamentally changed}.
Figure~\ref{fig:dropout_returns} suggests this point is somewhere between $0.1$ and $0.2$ for $p_\text{infer}$, though we suspect in practice this will be highly dependent on network architecture and the environment.

In Figure~\ref{fig:dropout_returns} we observe relatively weak returns on the real CarRacingFixedN environment when the inference dropout rate is zero.
Recall from Table~\ref{tab:rnn_loss} that the dropout variant has a much higher test loss than the non-dropout variant on CarRacingFixedN.
This means that when $p_\text{infer}=0$, the \textit{single} environment DDL is able to create is relatively inaccurate.
It is easier for the controller to exploit
any discrepancies between the dream environment and target environment because only a single dream environment exists.
However, as we increase the inference dropout rate it becomes harder for the controller to exploit the dynamics model,
suggesting that DDL is especially useful when it is difficult to learn an accurate World Model.

\subsection{When Should Dropout Masks be Randomized During Controller Training?}\label{ssec:dropout_abl}

In this ablation study we evaluate when the dropout mask should be randomized during training of $C$.
We consider two possible approaches of when to randomize the masks.
The first case only randomizes the mask at the beginning of an episode ({\em episode randomization\/}).
The second case samples a new dropout mask at every step
({\em step randomization\/}).
We also consider if it is effective to only apply dropout at inference time but not during $M$ training (i.e., $p_\text{infer}>0, p_\text{train}=0$).

As can be seen in Table~\ref{tab:mask_abl},
randomizing the mask at each step offers better returns on both target environments.
Better returns in the target environment when applying step randomization comes from the fact that the controller is exposed to a much larger number ($>1000\times$) of dream environments.
We also observe that applying step randomization without training the dynamics model with dropout yields a weak policy on the target environment.
This is due to the randomization fundamentally changing the task.
Training the dynamics model with dropout ensures that at inference time the masked model ($M^j$) is meaningful.

\begin{table}[ht]
    \centering
    \caption{Returns of the controller with different frequencies to randomize the dropout mask.}
    \begin{tabular}{ccc}
        \toprule
        & DoomTakeCover & CarRacingFixedN \\
        \midrule
        \makecell{episode randomization \\ ($p_\text{train}=0.05$, $p_\text{infer}=0.1$)} & $786 \pm 469$ & $601 \pm 197$ \\
        \makecell{step randomization \\ ($p_\text{train}=0.05$, $p_\text{infer}=0.1$)} & $ \mathbf{933 \pm 552}$ & $\mathbf{625 \pm 289}$ \\
        \makecell{step randomization \\ ($p_\text{train}=0$, $p_\text{infer}=0.1$)} & $339 \pm 90$ & $-43 \pm 52$ \\
        \bottomrule
    \end{tabular}
    \label{tab:mask_abl}
\end{table}
\vspace{-0.28cm}

\subsection{Comparison to Standard Regularization Methods}
In this experiment we compare Dropout's Dream Land with standard regularization methods.
First, we consider applying the standard use case of dropout ($0<p_\text{train}<1$ and $p_\text{infer}=0$).
Second, we consider a noisy variant of $M$ when training $C$.
The Noisy World Model uses exactly the same parameters for $M$ as the baseline World Model.
When training the controller, a small amount of Gaussian noise is added to $z$ at every step.

In Table~\ref{tab:car_reg},
we observe that DDL is better at generalizing from the dream environment to the target environment than the standard regularization methods.
Dropout World Models can be viewed as a regularizer on $M$.
Noisy World Models can be viewed as a regularizer on the controller $C$.
The strong returns on the target environment by DDL suggest that it is better at crossing the Dream2Real gap than standard regularization techniques.

\begin{table}[H]
    \centering
    \caption{Returns from World Models, Dropout World Models ($p_\text{train}=0.05$ and $p_\text{infer}=0.0$), Noisy World Models, and DDL ($p_\text{train}=0.05$ and $p_\text{infer}=0.1$) on the CarRacingFixedN and the original CarRacing environments.}
    \begin{tabular}{ccc}
        \toprule
        & CarRacingFixedN & CarRacing \\
        \midrule
        World Models & $399 \pm 135$ & $388 \pm 157 $ \\
        Dropout World Models & $-36 \pm 19$ & $-36 \pm 20$ \\
        Noisy ($\mathcal{N}(0, 1)$) World Models & $147 \pm 121$ & $180 \pm 132$ \\
        Noisy ($\mathcal{N}(0, 10^{-2})$) World Models & $455 \pm 171$ & $442 \pm 171$ \\
        Dropout's Dream Land & $\mathbf{625 \pm 289}$ & $\mathbf{610 \pm 267}$ \\
        \bottomrule
    \end{tabular}
    \label{tab:car_reg}
\end{table}

\subsection{Comparison to Explicit Ensemble Methods}
In this experiment we compare Dropout's Dream Land with two
other approaches for randomizing the dynamics of the dream environment.
We consider using an explicit ensemble of a population of dynamics models.
Each environment in the population was trained on the same set of trajectories described at the beginning of Section~\ref{sec:exp} with a different initialization and different mini-batches.
With the population of World Models we train a controller with Step Randomization
and a controller with Episode Randomization.
Note that the training cost of dynamics models and RAM requirements at inference time scale linearly with the population size.
Due to the large computational cost we consider a population size of $2$.

In Table~\ref{tab:car_rand}, we observe that neither Population World Models (PWM) Step Randomization or Episode Randomization substantially close the Dream2Real gap.
Episode Randomization does not dramatically improve results because the controller is forced to understand the hidden state ($\mathbf{h}$) representation of every $M$ in the population.
Step Randomization performs even worse than Episode Randomization because on top of the previously stated limitations, each dynamics model in the population is also forced to be compatible with the hidden state ($\mathbf{h}$) representation of all other dynamics models in the population.
DDL does not suffer from any of the previously stated issues and is also computationally cheaper because only one $M$ must be trained as opposed to an entire population. 

\begin{table}[H]
    \caption{Returns from World Models, PWM Episode Randomization, PWM Step Randomization, and DDL ($p_\text{train}=0.05$ and $p_\text{infer}=0.1$) on the CarRacingFixedN and the original CarRacing environments.}
    \centering
    \begin{tabular}{cccc}
        \toprule
        & CarRacingFixedN & CarRacing \\
        \midrule
        World Models & $399 \pm 135$ & $388 \pm 157 $ \\ 
        PWM Episode Randomization & $398 \pm 126$ & $402 \pm 142$ \\ 
        PWM Step Randomization & $-78 \pm 14$ & $-77 \pm 13$ \\ 
        Dropout's Dream Land & $\mathbf{625 \pm 289}$ & $\mathbf{610 \pm 267}$ \\
        \bottomrule
    \end{tabular}
    \label{tab:car_rand}
\end{table}

\section{Conclusion}
Dropout's Dream Land introduces a novel technique to improve controller generalization from dream environments to reality.
This is accomplished by taking inspiration from Domain Randomization and training the controller on a large set of different simulators.
A large set of different simulators are generated at little cost by the insight that dropout can be used to generate an ensemble of neural networks.
To the best of our knowledge this is the first work to bridge the reality gap between learned simulators and reality.
Previous work from Domain Randomization~\cite{tobin2017domain} is not applicable to learned simulators because they often do not have easily configurable parameters.
Future direction for this work could be modifying the dynamics model parameters 
in a targeted manner~\cite{wang2019paired,wang2020enhanced,such2019generative}.
This simple approach to generating different versions of a model could also be useful in committee-based methods~\cite{settles2009active,sekar2020planning}.

\clearpage
\bibliography{main}
\bibliographystyle{splncs04}

\clearpage

\appendix
\section{Appendix}

\subsection{Architecture Details} \label{A:arch}
We follow the same architecture setup as World Models.
We adopt the following notation~\cite{kim2020learning} to describe the VAE architecture.

\textbf{Conv2D(a, b, c)}: 2D-Convolution layer with output channel size \textbf{a}, kernel size \textbf{b}, and stride \textbf{c}.
All use valid padding and ReLU activations.

\textbf{T.Conv2D(a, b, c)}: Transposed 2D-Convolution layer with output channel size \textbf{a}, kernel size \textbf{b}, stride \textbf{c}.
The final layer uses a sigmoid activation but every other layer uses ReLU activations.

\textbf{LSTM(a)}: LSTM layer with \textbf{a} units.

\textbf{Dense(a)}: Fully Connected layer with output size \textbf{a} followed by a ReLU activation.

\textbf{Linear(a)}: Linear layer with output size \textbf{a}.

\textbf{Reshape(a)}: Reshape input to output size \textbf{a}.

\begin{table}[ht]
    \centering
    \caption{Architectures DoomTakeCover and CarRacing.}
    \begin{tabular}{c c}
        \toprule
        DoomTakeCover & CarRacing \\
        \toprule
        \multicolumn{2}{c}{VAE ($V$)} \\
        \midrule
        Conv2D(32, 4, 2) & Conv2D(32, 4, 2) \\
        Conv2D(64, 4, 2) & Conv2D(64, 4, 2) \\
        Conv2D(128, 4, 2) & Conv2D(128, 4, 2) \\
        Conv2D(256, 4, 2) & Conv2D(256, 4, 2) \\
        Reshape(1024) & Reshape(1024) \\
        Linear(32), Linear(32) & Linear(64), Linear(64) \\
        Dense(1024) & Dense(1024) \\
        Reshape(1, 1, 1024) & Reshape(1, 1, 1024) \\
        T.Conv2D(128, 5, 2) & T.Conv2D(128, 5, 2) \\
        T.Conv2D(64, 5, 2) & T.Conv2D(64, 5, 2) \\
        T.Conv2D(32, 6, 2) & T.Conv2D(32, 6, 2) \\
        T.Conv2D(3, 6, 2) & T.Conv2D(3, 6, 2) \\
        \midrule
        \multicolumn{2}{c}{World Model ($M$)} \\
        \midrule
        LSTM(512) & LSTM(256) \\
        Dense(961) & Dense(482) \\
        \midrule
        \multicolumn{2}{c}{Controller ($C$)} \\
        \midrule
        Linear(1) & Linear(3) \\
        \bottomrule
    \end{tabular}
    \label{tab:arch}
\end{table}

\end{document}